\documentclass[10pt,twocolumn,letterpaper]{article}

\usepackage{iccv}
\usepackage{times}
\usepackage{epsfig}
\usepackage{graphicx}
\usepackage{amsmath}
\usepackage{amssymb}
\usepackage{bm}
\usepackage{booktabs}
\usepackage[dvipsnames,table,xcdra]{xcolor}
\usepackage{multirow}
\usepackage{pifont}
\usepackage{appendix}

\usepackage[breaklinks=true,colorlinks,bookmarks=false,citecolor=RoyalBlue]{hyperref}

\makeatletter\renewcommand\paragraph{\@startsection{paragraph}{4}{\z@}
  {0.4em \@plus.0ex \@minus.2ex}{-.5em}{\normalfont\normalsize\bfseries}}\makeatother

\iccvfinalcopy 

\def\iccvPaperID{3683}

\newcommand{\OURS}{UniVRD\xspace}
\newcommand{\p}[1]{\paragraph{#1:}}
\newcommand{\authorskip}{\quad}

\newcommand{\cmark}{\ding{51}}
\newcommand{\xmark}{\ding{55}}
\DeclareMathOperator*{\argmax}{arg\,max}

\newcommand{\rowA}{\rowcolor{gray!50}}
\newcommand{\rowB}{\rowcolor{gray!15}}
\newcommand{\celG}[1]{\textcolor{gray!90}{#1}}
\newcommand{\numD}[1]{\textcolor{red}{$_{\mathbf{{\downarrow #1}}}$}}
\newcommand{\numU}[1]{\textcolor{blue}{$_{\mathbf{{\uparrow #1}}}$}}

\begin{document}

\title{Unified Visual Relationship Detection with Vision and Language Models}

\author{Long Zhao \authorskip Liangzhe Yuan \authorskip Boqing Gong \authorskip Yin Cui \authorskip Florian Schroff\\ Ming-Hsuan Yang \authorskip Hartwig Adam \authorskip Ting Liu\\[2.6mm]
Google Research\\{\tt\small \{longzh,liuti\}@google.com}}

\maketitle

\begin{abstract}
This work focuses on training a single visual relationship detector predicting over the union of label spaces from multiple datasets. Merging labels spanning different datasets could be challenging due to inconsistent taxonomies. The issue is exacerbated in visual relationship detection when second-order visual semantics are introduced between pairs of objects. To address this challenge, we propose UniVRD, a novel bottom-up method for \textbf{Uni}fied \textbf{V}isual \textbf{R}elationship \textbf{D}etection by leveraging vision and language models (VLMs). VLMs provide well-aligned image and text embeddings, where similar relationships are optimized to be close to each other for semantic unification. Our bottom-up design enables the model to enjoy the benefit of training with both object detection and visual relationship datasets. Empirical results on both human-object interaction detection and scene-graph generation demonstrate the competitive performance of our model. UniVRD achieves 38.07 mAP on HICO-DET, outperforming the current best bottom-up HOI detector by 14.26 mAP. More importantly, we show that our unified detector performs as well as dataset-specific models in mAP, and achieves further improvements when we scale up the model. Our code will be made publicly available on GitHub\footnote{\url{https://github.com/google-research/scenic/tree/main/scenic/projects/univrd}}.
\end{abstract} 
\section{Introduction}
\label{sec:introduction}

\begin{figure}[t]
\begin{center}
\includegraphics[width=\linewidth]{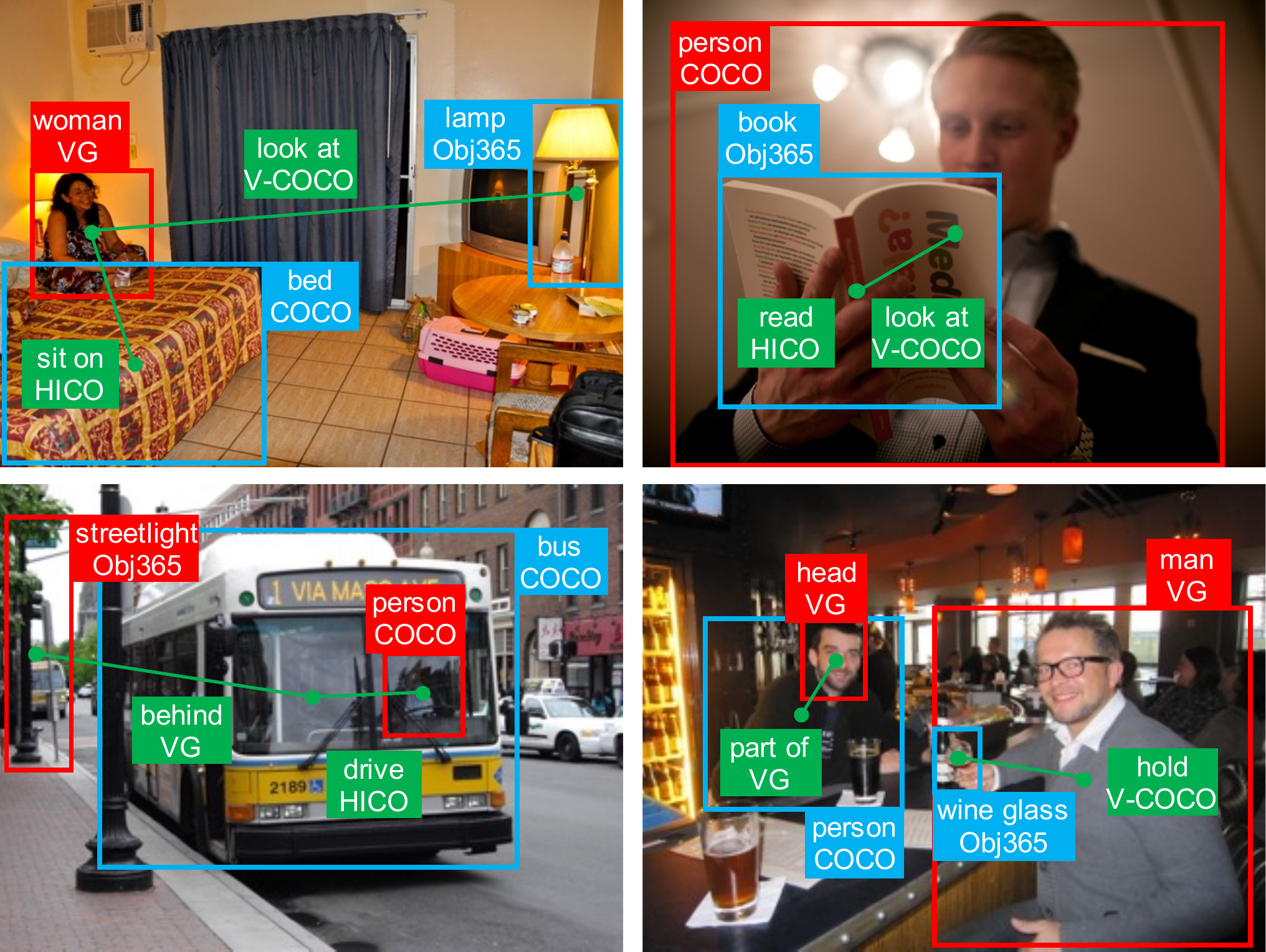}
\end{center}
\vspace{-0.6mm}
\caption{Different VRD datasets provide different sets of unary object classes and binary relationships. We train a single visual relationship detector to unify their label spaces that generalizes across datasets. For each visual relationship, we highlight its subject in red, predicate in green, and object in blue.}
\label{fig:teaser}
\end{figure}

Visual relationship detection (VRD) is a fundamental problem in computer vision, where visual relationships are typically defined over pairs of localized objects, connected with a predicate.
Despite the availability of a diverse set of data with rich pair-wise object annotations~\cite{chao2018learning,gupta2015visual,hudson2019gqa,krishna2017visual}, existing VRD models, however, are typically focusing on training from a single data source.
The resultant models are therefore restricted in both image domains and text vocabularies, limiting their generalization and scalability. 
\textit{Can we train a single visual relationship detector that unifies diverse datasets with heterogeneous label spaces?}

Consider Figure~\ref{fig:teaser}, labels for objects and relations across datasets are non-disjoint and therefore could be synonymous (\eg, `read' in HICO-DET~\cite{chao2018learning} \vs `look at' in V-COCO~\cite{gupta2015visual}), subsidiary (\eg, `person' in COCO~\cite{lin2014microsoft} \vs `woman' / `man' in Visual Genome~\cite{krishna2017visual}), or overlapping (\eg, `wine glass' in Objects365~\cite{shao2019objects365} \vs `glass' in Visual Genome). Furthermore, VRD models need to infer relationships (\ie, predicates) of second-order visual semantics between objects. The combinatorial complexity elevates the challenge to a new level. Depends on context, the same object or predicate might appear in different tenses or forms (\eg, `man' \vs `men' or `wears' \vs `wearing' in Visual Genome) and their meanings may vary (\eg, `eating a sandwich' \vs `eating (with) a fork' in V-COCO). Therefore, manually curating a merged label space spanning different datasets for training a unified VRD model is difficult.

On the other hand, recent breakthroughs in vision and language models (VLMs) that are jointly trained on web-scale image-text pairs (\eg, CLIP~\cite{radford2021learning} and ALIGN~\cite{jia2021scaling}) provide an alternative direction to approach our challenge.
Intuitively, benefiting from the large language encoders~\cite{devlin2019bert,raffel2020exploring} and contrastive image-text co-training, a pre-trained VLM should be able to encode ``similar visual relationships'' close to each other in the embedding space. These relationships contain similar action, subject, and object labels in semantics, \eg, ``a person watching a television'' \vs ``a man looking at a TV''. They are commonly measured by distances between semantic words or language embeddings~\cite{lu2016visual}, which motivates us to use large language models for unification.
Specifically, the learnt text embeddings of VLMs can be used to reconcile heterogeneous label spaces across VRD datasets of similar visual relationships, while their jointly trained image encoders ensure the alignment with the visual content.

In light of this, we propose \textit{\OURS} (\textbf{Uni}fied \textbf{V}isual \textbf{R}elationship \textbf{D}etection), a bottom-up framework consisting of an object detector and pair-wise relationship decoder in a cascaded manner. To fine-tune VLMs for object detection, we adopt an encoder-only architecture~\cite{minderer2022simple} and attach a minimal set of heads to each Transformer output token so that the learnt knowledge from the image-level pre-training can be preserved. A lightweight Transformer decoder~\cite{alayrac2022flamingo} is then appended to the object detector for decoding pair-wise relationships from the predicted objects by formulating the optimization target as a set prediction problem~\cite{carion2020end}. Further, we use natural languages in place of categorical integers to define and unify the label space. Our bottom-up design and language-defined label space enable the model to enjoy various existing object detection and visual relationship detection datasets for training, yielding strong scalability and substantial performance improvements over existing bottom-up detection approaches.

We evaluate our approach on two VRD tasks: human-object interaction (HOI) detection and scene graph generation (SGG). Crucially, we demonstrate competitive performances on both tasks --- our model achieves the state of the art on HICO-DET ($38.07$ mAP), a substantial improvement of $14.26$ mAP over the current best-performing bottom-up HOI detector. For the first time, we show that a unified model can perform as well as dataset-specific ones, and obtain notable improvements in mAP on long-tailed VRD datasets when the model is scaled up.

In summary, this paper makes the following \textbf{main contributions}: (1) a novel VRD framework that unifies multiple datasets which cannot be done by previous work without VLMs; (2) an effective and strong model training recipe, including improvements on models, losses, augmentations, training paradigms, \etc; (3) state-of-the-art results showing strong scalability and generalization of our model. Our design is simple, interoperable, and can easily leverage new advances in VLMs. We believe our work is first-of-its-kind that brings new insights to the community and as a flexible starting point for future research on tasks requiring visual relationship understanding. \section{Related Work}
\label{sec:related_work}

\p{Visual relationship detection} Visual relationship detection/prediction (VRD) is first proposed in~\cite{lu2016visual} then formulated as a dual-graph generation task called scene graph generation (SGG) by~\cite{xu2017scene}. Prior methods~\cite{li2021bipartite,xu2017scene} refine the representations in a scene graph by a message passing mechanism. More recent works aim to eliminate data bias in SGG (\ie, unbiased/informative SGG) during the inference process by using graph semantic relationships~\cite{yu2020cogtree}, bi-level resampling strategies~\cite{li2021bipartite}, or data augmentation~\cite{abdelkarim2021exploring}.

Human-object interaction (HOI) detection~\cite{chao2018learning}, as a popular VRD task, aims to detect human-object pairs and infers their interactions. Existing HOI detectors can be summarized into two paradigms: bottom-up methods~\cite{gao2020drg,gao2018ican,gkioxari2018detecting,qi2018learning,wan2019pose,wang2020contextual} and single-stage methods~\cite{cong2023reltr,kim2021hotr,liao2022gen,tamura2021qpic,zhang2021mining,zou2021end}. Bottom-up methods detect instances first and predict interactions based on them, while single-stage methods detect all HOI triplets directly and simultaneously in an end-to-end manner. Our method uses a bottom-up design and outperforms both types of methods thanks to label unification and knowledge transferred from image-level pre-training.

Some prior works~\cite{gao2020drg,kim2020detecting,liu2020amplifying,zhong2020polysemy} incorporate language priors~\cite{lu2016visual} to VRD based on the observation that relationships are semantically related to each other in the language space. They use word embeddings~\cite{mikolov2013efficient} to cast relationships into a vector space so that similar visual relationships are close to each other. However, they are limited to a small and fixed set of semantic categories. Different from them, we use well-aligned image-text embeddings to capture semantic relationships in VRD, which are more powerful.

\p{Unifying label spaces from multiple datasets} Training on multiple datasets is an effective way to improve model generalization. Prior works~\cite{hasan2021generalizable,ranftl2020towards,yang2019hierarchical,zhao2020object,zhou2022simple} focus on merging different visual semantic concepts across different label spaces for object detection or segmentation tasks. MSeg~\cite{lambert2020mseg} manually unifies the taxonomies of different semantic segmentation datasets and utilizes Amazon Mechanical Turk to resolve inconsistent annotations. Zhou~\etal~\cite{zhou2022simple} propose to learn a label space from visual data automatically for object detection, without requiring any manual effort. Zhao~\etal~\cite{zhao2020object} train a universal object detector by manually merging the taxonomies and generating cross-dataset pseudo-labels. Most methods~\cite{lambert2020mseg,wang2019towards,zhao2020object} find a performance drop when training a single unified model. Instead, we train a unified VRD model and show that it can perform as well as dataset-specific ones in mAP.  

\p{Vision and language models} Over the past several years, there are a surge of works that leverage vision and language models (VLMs) to build vision systems~\cite{jia2021scaling,li2021align,radford2021learning,yu2022coca,zhai2022lit}. By using large amounts of image-text data for model training, VLMs are able to learn well-aligned image and text embeddings, yielding major improvements in open-vocabulary and zero-shot classification tasks~\cite{li2022grounded,yuan2022rlip,zareian2021open,zhao2022exploiting}. Some recent studies~\cite{yao2022pevl,yao2021cpt,yuan2022rlip} propose to employ VLMs for VRD. They focus on designing efficient pretext tasks to capture object relationships and show promising few-shot and transfer learning ability. In contrast, we take the advantage of image-text embeddings from pre-trained VLMs~\cite{radford2021learning,zhai2022lit} for label space unification, which has never been explored. \begin{figure*}[t]
\begin{center}
\includegraphics[width=\linewidth]{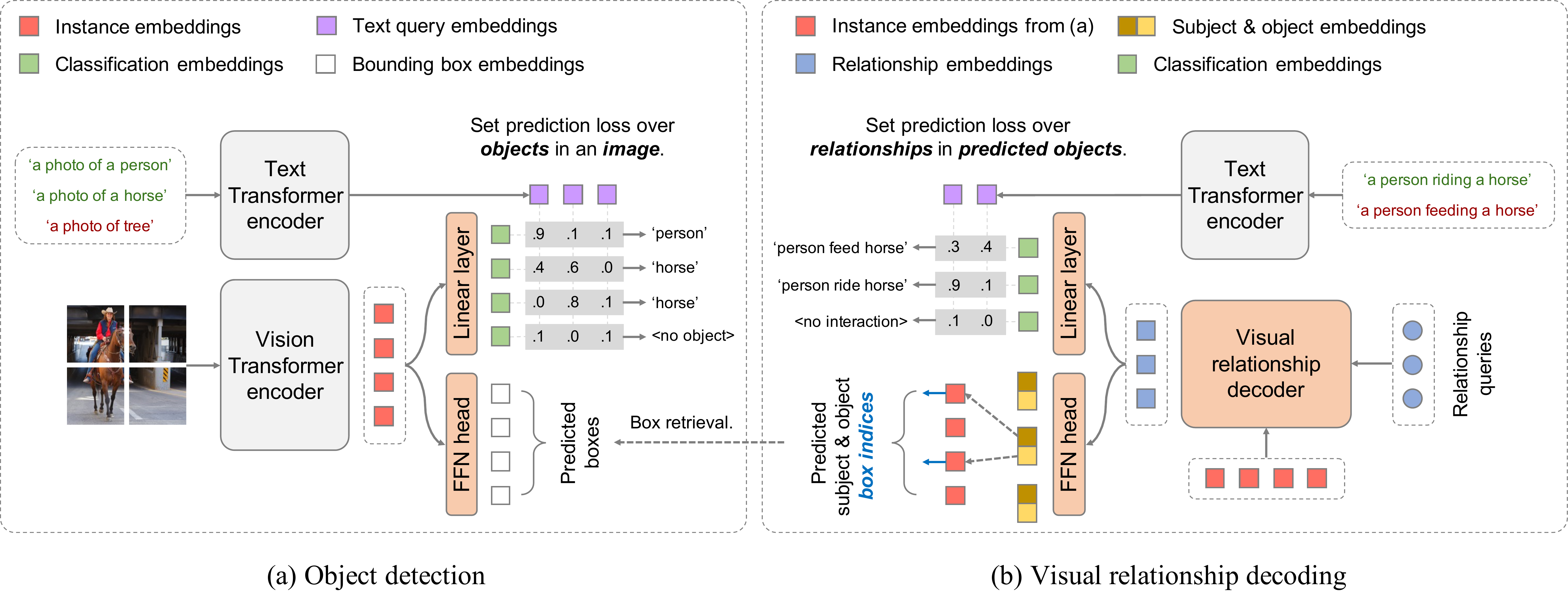}
\end{center}
\vspace{-2.6mm}
\caption{\textbf{Overview of our method.} \textbf{(a)} We first adapt a pre-trained vision and language model (VLM) for object detection. \textbf{(b)} We then append a decoder to the output instance embeddings for decoding pair-wise relationships from predicted objects. Query strings are embedded with the text encoder and used for classification in the unified language space. The pre-trained VLM encoders are marked in light gray and our attached modules are in light orange. Positive and negative text strings are highlighted in green and red, respectively.}
\label{fig:pipeline}
\end{figure*}

\section{Approach}
\label{sec:approach}

The goal of VRD is to predict a set of $\langle$subject, predicate, object$\rangle$ triplets, representing the bounding box of a subject, that of an object, and multi-label relationship types, from a given image. The proposed \OURS presents a bottom-up recipe: (1) adapting VLM for object detection by adding detection heads; (2) decoding visual relationships among the detected objects through their fine-tuned visual embeddings. Each of them is viewed as a direct \emph{set prediction problem}~\cite{carion2020end}: we use Hungarian algorithm~\cite{kuhn1955hungarian} to find a bipartite matching between ground-truth and prediction, and compute losses only on matched pairs. The trained model can be queried in different ways (\eg, natural languages or visual embeddings) to perform relationship detection. Our full pipeline is illustrated in Figure~\ref{fig:pipeline}. 

We consider training \OURS on multiple datasets. After merging them, we have two label spaces: $\mathcal{C}_\text{obj}$ for objects and $\mathcal{C}_\text{rel}$ for relationships. Both of them can be ambiguous as discussed in Section~\ref{sec:introduction}, and we convert them to language spaces $\mathcal{C}_\text{obj} \rightarrow \mathcal{T}_\text{obj}, \mathcal{C}_\text{rel} \rightarrow \mathcal{T}_\text{rel}$ as shown later. We allow training with datasets containing no relationships and they downgrade to object detection datasets. Next, we describe the bottom-up recipe in the following sections.

\subsection{Architecture}

\p{Object detector} Our model uses a standard Vision Transformer (ViT)~\cite{dosovitskiy2020image} as the image encoder and a similar architecture as the text encoder, which is a common configuration in two-tower VLMs~\cite{jia2021scaling,radford2021learning,zhai2022lit}. To adapt the image encoder for object detection, we fine-tune the model by predicting one object instance directly from each image token. We remove the pooling and final projection layers, and instead project each output token representation (\ie, instance embedding\footnote{An instance refers to an object instance in the image.}) to get the per-instance classification embedding by a linear layer. Bounding box coordinates are obtained by passing instance embeddings through a feed-forward network (FFN). The final outcomes of the detector are a set of predicted bounding boxes $\mathcal{B} = \{\bm{b}_i\}^N_{i=1}$ and their corresponding instance embeddings $\mathcal{Z} = \{\bm{z}_i\}^N_{i=1}$, where $N$ is the maximum number of predicted objects and it equals to the number of tokens (\ie, sequence length) of the image encoder. 

This encoder-only design resembles DETR~\cite{carion2020end}, but is simplified by removing the decoder, leading to several advantages. First, it ensures that nearly all of the parameters (of both image and text encoders) can benefit from image-level pre-training, without the need for knowledge distillation~\cite{gu2021open} or detection-tailored pre-training~\cite{zhong2022regionclip}. Second, the fine-tuned instance embeddings can be directly utilized for visual relationship decoding without feature pooling in conventional bottom-up methods~\cite{gao2018ican,gkioxari2018detecting,qi2018learning}, which further reduces the model complexity.

\p{Relationship decoder} We append a Transformer decoder to the object detector for decoding visual relationships from its output. In similar spirit to query-based models~\cite{carion2020end,jaegle2021perceiver}, we learn a pre-defined number of latent input queries, \ie, \textit{relation queries}. The decoder then takes as input a set of relation queries and instance embeddings $\mathcal{Z}$ predicted by the object detector. These relation queries are fed to a Transformer stack that attends to the instance embeddings to produce relation embeddings $\mathcal{R} = \{\bm{r}_j\}^M_{j=1}$, where $M$ is the number of output relation embeddings from the decoder, equal to the number of learnt relation queries. Additionally, the keys and values computed from the learnt latents are concatenated to the keys and values obtained from $\mathcal{Z}$ like Perceiver Resampler~\cite{alayrac2022flamingo}, which has been proven more efficient than a plain Transformer decoder. We then apply one linear layer and one FFN on relation embeddings to predict per-relationship embeddings for classification and locations of subject and object boxes, respectively.

In contrast to single-stage methods~\cite{liao2022gen,tamura2021qpic,zhang2021mining}, we let the model predict the indices of bounding boxes outputted by the object detector instead of box coordinates. This design avoids making redundant predictions for the same instance, which improves the model efficiency. Our model finds the indices of subject and object boxes through comparing the predicted relation embeddings $\bm{r} \in \mathcal{R}$ with the instance embeddings $\bm{z} \in \mathcal{Z}$. To be specific, we project each relation embedding $\bm{r}_j$ using a FFN into a subject embedding $f_\text{sub}(\bm{r}_j)$ and an object embedding $f_\text{obj}(\bm{r}_j)$. The subject index $s_j$ and object index $o_j$ are obtained by:
\begin{equation}
\label{eq:indices}
\begin{aligned}
s_j &= \argmax_{\bm{z} \in \mathcal{Z}} \left\{\text{sim}\left(f_\text{sub}(\bm{r}_j), \bm{z}\right)\right\},\\
o_j &= \argmax_{\bm{z} \in \mathcal{Z}} \left\{\text{sim}\left(f_\text{obj}(\bm{r}_j), \bm{z}\right)\right\},
\end{aligned}
\end{equation}
where $\text{sim}(\cdot, \cdot)$ measures the cosine similarity between two embeddings. We can then retrieve the corresponding subject box $\bm{b}_{s_j}$ and object box $\bm{b}_{o_j}$ from $\mathcal{B}$.

\p{Text embeddings for classification} We use text embeddings, rather than class integers, to classify detected objects. The text embeddings, also called \textit{text queries}, are obtained by converting category names or textual object descriptions (\eg, `person') to prompts (\eg, `a photo of a person')~\cite{rao2022denseclip} and passing them through the text encoder. For each object, the task of the detector then becomes to predict a bounding box and a class probability over text queries. We note that text queries can be different for each image. In effect, all images therefore have a shared discriminative label space $\mathcal{T}_\text{obj}$, which is defined by a set of text strings. Note that we do not add a `background' class to the label space like traditional detectors~\cite{carion2020end,tamura2021qpic}, because this avoids imposing penalties on positive samples not exhaustively annotated, which commonly occur in merged datasets~\cite{zhao2020object}.

To classify detected relationships, we use text queries in a similar way as classifying objects in our object detector. The difference is that their label space $\mathcal{T}_\text{rel}$ is defined by pair-wise relationship triplets, rather than unary instance category names. To this end, we cast a set of relationship triplets $\langle$subject, predicate, object$\rangle$ (\eg, $\langle$person, ride, horse$\rangle$) into prompts (\eg, `a person riding a horse') and feed them into the text encoder to get text queries.

\subsection{Data Augmentation}

\p{Mosaics} We apply `mosaics' image augmentation technique for training both our object detector and relationship decoder. It aims to increase the range of image scales seen by the model, which is achieved by assembling multiple images into grids of varying sizes: we randomly sample single images, $2 \times 2$ image grids, and $3 \times 3$ image grids, with probabilities of $0.4$, $0.3$, and $0.3$, respectively. Using `mosaics' image augmentation provides two major benefits to model training. First, this procedure allows us to use widely varying image scales while avoiding excessive padding and the need for variable model input size during training. Second, it effectively fuses samples from object detection and visual relationship detection datasets within a batch, when our model is trained in an end-to-end manner.

\p{Text prompting} When generating text queries for object categories and relationship triplets, we augment the input text strings using prompt templates. To handle object categories, we use the prompt templates proposed by CLIP~\cite{radford2021learning} (such as `a photo of a $\langle$object$\rangle$', where $\langle$object$\rangle$ is replaced by the category name). When training the object detector, we randomly sample from the $80$ CLIP prompt templates to ensure that, within an image, every instance of a category has the same prompt, but prompt templates differ between categories and across images. This sampling technique largely reduces the number of text embeddings needed to be computed within each batch, which
reduces the training time and memory cost.

We produce text queries for relationship triplets in a similar way as prompting object names. The major difference is that we use a single prompt template (\ie, `a $\langle$subject$\rangle$ $\langle$predicate$\rangle$-ing a $\langle$object$\rangle$', where $\langle$subject$\rangle$ and $\langle$object$\rangle$ are replaced by subject and object categories, respectively; $\langle$predicate$\rangle$-ing\footnote{``$\langle$predicate$\rangle$-ing'' is obtained by using a standard Python NLP library: \url{https://github.com/clips/pattern}.} is the present continuous tense of the predicate) to prompt relationships. In addition, we use the word `and' to represent the `no-interaction' predicate category contained in some datasets (\eg, HICO-DET~\cite{chao2018learning}). The same prompt template is used for inference and we do not apply prompt ensemble~\cite{minderer2022simple,radford2021learning}, as no performance improvements are observed if CLIP-like prompt templates are employed. This is potentially because unlike object categories, each of which is usually a single word, relationship triples are combinations of subjects, predicates, and objects, which already capture rich semantic contents for the text encoder to generate meaningful language embeddings.

\subsection{Model Training}

\p{Training strategy} Our whole network can be trained in either an end-to-end fashion or multiple stages. Empirically, we found the direct end-to-end training of the whole network from scratch does not work well, likely because of the dependency between the two modules and highly non-linear property of the bipartite matching losses. Thus, we propose a cascade training paradigm that we found is more stable and effective in practice. Concretely, in the first stage, we initialize the object detector using images with bounding box annotations. In the second stage, we train the visual relationship decoder, where images with relationship annotations are used. Furthermore, we found whether to freeze or fine-tune the object detector in the second stage highly depends on the scale of the training data. When the training data are limited, a frozen object detector prevents overfitting; otherwise, fine-tuning it leads to notable performance improvements if more training data are available.

\p{Loss functions} The training objective of our model is similar to DETR~\cite{carion2020end} by using the bipartite matching loss, but we adapt it for training bottom-up VRD models as follows.

To train the object detector, we use the ground-truth object category names and sampled negatives as text queries for each image. Negatives are randomly sampled categories in proportion to their frequency in the data from the unified label space, and we have at least 50 negatives per image~\cite{zhou2021probabilistic}. The classification head then outputs logits over the per-image label space $\mathcal{T}_\text{obj}' \subseteq \mathcal{T}_\text{obj}$ defined by the text queries. To be specific, we let $f_\text{cls}(\bm{z})$ be the class embedding projected from the instance embedding $\bm{z}$ using a linear layer $f_\text{cls}$, and $\bm{t}_i$ be a text query from the label space $\mathcal{T}_\text{obj}'$. The classification loss can be written as:
\begin{equation}
\label{eq:cls_loss}
\begin{gathered}
\mathcal{L}_\text{cls}(\bm{z}, \mathcal{T}_\text{obj}'; \hat{\bm{y}}) = \mathcal{L}_\text{CE}(\bm{e}, \hat{\bm{y}})\\
\text{and} \; \bm{e} = \left[\text{sim}(f_\text{cls}(\bm{z}), \bm{t}_1), \cdots, \text{sim}(f_\text{cls}(\bm{z}), \bm{t}_{|\mathcal{T}_\text{obj}'|})\right],
\end{gathered}
\end{equation}
where $\hat{\bm{y}}$ denotes the multi-hot ground-truth label. $\mathcal{L}_\text{CE}$ is the cross-entropy loss: $\mathcal{L}_\text{CE} = -\sum_{i} \hat{y}_i\log(p_i)$ and $p_i = \exp(e_i/\tau)/\sum_i \exp(e_i/\tau$), where $\tau$ is a learnable temperature, and $\hat{y}_i$ denotes the $i$-th element in $\hat{\bm{y}}$, as $e_i$ does in $\bm{e}$. In practice, $\mathcal{L}_\text{CE}$ in Eq.~\eqref{eq:cls_loss} is replaced by the \textit{focal sigmoid cross-entropy loss}~\cite{lin2017focal,zhu2020deformable}, since the training datasets have non-disjoint label spaces. We use the box loss $\mathcal{L}_\text{box}$~\cite{carion2020end}, \ie, a linear combination of the $\ell_1$ loss and the generalized IoU loss~\cite{rezatofighi2019generalized}, to train the box regression head $f_\text{box}$ by optimizing the difference between the predictions $\bm{b} = f_\text{box}(\bm{z})$ and ground-truth box coordinates $\hat{\bm{b}}$. Then the Hungarian loss for our object detector is defined as:
\begin{equation}
\label{eq:od_loss}
\mathcal{L}_\text{OD} = \frac{1}{N} \sum_{i=1}^N \mathcal{L}_\text{cls}(\bm{z}_i, \mathcal{T}_\text{obj}'; \hat{\bm{y}}_i) + \mathcal{L}_\text{box}(\bm{b}_i; \hat{\bm{b}}_i),
\end{equation}
where $N$ is the number of image tokens.

We use a bipartite matching loss similar to Eq.~\eqref{eq:od_loss} for optimizing the relationship decoder, with two major modifications. First, since the model predicts box indices rather than coordinates, the box loss $\mathcal{L}_\text{box}$ in Eq.~\eqref{eq:od_loss} is re-formulated into an index prediction loss $\mathcal{L}_\text{ind}$. In particular, given a ground-truth relationship containing a pair of subject and object boxes, we let their corresponding ground-truth one-hot subject index $\hat{\bm{s}}$ and object index $\hat{\bm{o}}$ be the indices of their best-matching predicted boxes generated by the object detector. Let its matched relation embedding be $\bm{r}$, we further define the index prediction loss as:
\begin{equation}
\label{eq:ind_loss}
\mathcal{L}_\text{ind}(\bm{r}; \hat{\bm{s}}, \hat{\bm{o}}) = \mathcal{L}_\text{cls}'(\bm{r}, \mathcal{Z}; \hat{\bm{s}}) + \mathcal{L}_\text{cls}'(\bm{r}, \mathcal{Z}; \hat{\bm{o}}),
\end{equation}
where $\mathcal{Z}$ denotes a set of instance embeddings calculated from the input image; $\mathcal{L}_\text{cls}'$ is a variant of $\mathcal{L}_\text{cls}$ in Eq.~\eqref{eq:cls_loss} with $\mathcal{L}_\text{CE}$ replaced by the \textit{focal softmax cross-entropy loss}, since both $\hat{\bm{s}}$ and $\hat{\bm{o}}$ are one-hot. Second, we replace sampled object text queries $\mathcal{T}_\text{obj}'$ by relationship text queries $\mathcal{T}_\text{rel}' \subseteq \mathcal{T}_\text{rel}$ to classify visual relationships. The final Hungarian loss for visual relationship decoding can be written as:
\begin{equation}
\label{eq:vrd_loss}
\mathcal{L}_\text{VRD} = \frac{1}{M} \sum_{j=1}^M \mathcal{L}_\text{cls}(\bm{r}_j, \mathcal{T}_\text{rel}'; \hat{\bm{c}}_j) + \mathcal{L}_\text{ind}(\bm{r}_j; \hat{\bm{s}}_j, \hat{\bm{o}}_j),
\end{equation}
where $\hat{\bm{c}}_j$ denotes the multi-hot ground-truth label for the $j$-th predicted relationship. Note that our box index prediction loss shares a similar concept with the \textit{HO Pointers}~\cite{kim2021hotr}, but differs in the model design and loss computation.

\subsection{Inference}
\label{sec:approach:inference}

Our inference pipeline assembles the outputs of the object detector and relationship decoder to form relationship triplets. Formally, given an input image, the object detector predicts a set of object boxes $\{\bm{b}_i\}^N_{i=1}$, and the relation decoder generates relation embeddings $\{\bm{r}_j\}^M_{j=1}$ together with their relevant subject indices $\{s_j\}^M_{j=1}$ and object indices $\{o_j\}^M_{j=1}$. We combine them to obtain relationship triplets $\{\langle \bm{b}_{s_j}, \bm{b}_{o_j}, \bm{r}_j\ \rangle\}^M_{j=1}$ via box retrieval (\ie, using the box index to get the corresponding box coordinates). Then, given a text query embedding $\bm{t}$ representing a relationship string (\eg, `a person riding a horse'), the triplet score is computed by $\text{sim}(\bm{r}_j, \bm{t})$. Among the top-$K$ scored triplets, we perform pair-wise non-maximum suppression (PNMS)~\cite{zhang2021mining} within each relationship class, namely \textit{per-class PNMS}, to filter out highly-overlapping results.

\p{One-shot transfer} We note that the proposed model does not require query embeddings to be of textual origin, which means we can provide image- instead of text-derived embeddings as queries to the classification head without modifying or re-training the model. By leveraging embeddings of prototypical visual relationships as queries, our model can therefore perform image-conditioned relationship detection. This allows detection of visual relationships which might be hard to describe in text. \section{Experiments}
\label{sec:experiments}

The proposed method is evaluated on two popular VRD tasks: human-object interaction (HOI) detection and scene graph generation (SGG). We experiment with two configurations: training a single model with each individual VRD dataset (dataset-specific ones in Sections~\ref{sec:experiments:hoi} and \ref{sec:experiments:sgg}) and multiple VRD datasets (unified ones in Section~\ref{sec:experiments:unified}). We analyze model performance and scalability under both configurations and improvements obtained by a unified model.

We note that our method is the first to use VLMs to unify multiple datasets for VRD. There are no such configurations in previous work, so making a perfectly controlled comparison on the pre-trained strategy and data is infeasible. Therefore, we perform \emph{system-level} comparisons instead, where our goal is to situate our method in the context of current state-of-the-art methods.

\subsection{Experimental Settings}

\p{Datasets} For HOI detection tasks, we conduct experiments on HICO-DET~\cite{chao2018learning} and
V-COCO~\cite{gupta2015visual}. The HICO-DET~(HICO) dataset contains $37,536$ training images and $9,515$ test images, including $600$ HOI triplets derived from the combinations of $117$ verbs and $80$ objects. We evaluate under the \textit{Default} setting. V-COCO comprises $2,533$ training images, $2,876$ validation images, and $4,946$ test images. This benchmark is annotated with $24$ actions and $80$ objects. Note that the object classes in HICO-DET and V-COCO are identical to COCO~\cite{lin2014microsoft}.

We use Visual Genome (VG)~\cite{krishna2017visual} for SGG tasks. This dataset contains $108,077$ images annotated with free-form text for a wide array of objects and relationships ($100,298$ object annotations and $36,515$ relation annotations). We adopt the most common data splits from~\cite{xu2017scene} that remove rare categories by selecting the top-$150$ object categories and top-$50$ predicate categories by frequency. The entire dataset is then divided into training and test sets by the ratio of $7$ to $3$. To further improve the object detection accuracy of our model, we incorporate COCO~\cite{lin2014microsoft} and Objects365 (O365)~\cite{shao2019objects365} during training.

\p{Metrics} For both HOI detection and SGG tasks, models are required to first detect bounding boxes and then recognize object and predicate classes of a relationship. On the HICO dataset, we follow the default setting and report the mAP over three different category sets: all $600$ HOI categories in HICO (Full), $138$ HOI categories with fewer than $10$ training instances (Rare), and $462$ HOI categories with $10$ or more training instances (Non-Rare). The common evaluation metric on V-COCO is role AP, which ignores object classes. We instead report mAP on V-COCO to make the evaluation protocol consistent across datasets. In addition, we report mAP for two scenarios: Scenario \#1 includes cases even without any objects (for the four action categories of body motions, \eg, $\langle$person, walk$\rangle$), and Scenario \#2 ignores these cases. We follow the conventional mean Recall@$K$ ($K$ equals to $50$ or $100$) as the evaluation metrics~\cite{lu2016visual,xu2017scene,zellers2018neural} on VG and report mAP as well.

\p{Implementation details} The proposed method is implemented using JAX~\cite{jax2018github} and the \emph{Scenic} library~\cite{dehghani2022scenic}. All models are trained on TPUv3 hardware. We experiment with two VLMs: CLIP~\cite{radford2021learning} and LiT~\cite{zhai2022lit}. We apply random crop, random horizontal flip, and mosaics as image augmentations. The number of relation queries is set to $100$ and per-class PNMS threshold to $0.7$. The object detector is trained following the setup in~\cite{minderer2022simple} using HICO, COCO, O365, and VG, except that the text encoder is frozen. The relationship decoder is optimized by the Adam optimizer~\cite{kingma2015adam} with a learning rate of $1.0 \times 10^{-4}$ and $64$ batch size. We use cosine learning rate decay~\cite{loshchilov2017sgdr} and per-example global norm gradient clipping. Please refer to the supplementary material for more details.

\begin{table}[t]
\begin{center}
\resizebox{\linewidth}{!}{
\begin{tabular}{lcccc}
\toprule
& & \multicolumn{3}{c}{Default ($\%$)}\\
\cmidrule(lr){3-5}
Model & Extra-sup. & $\text{mAP}_\text{F}$ & $\text{mAP}_\text{R}$ & $\text{mAP}_\text{N}$\\
\midrule
\rowA \multicolumn{5}{l}{\textit{Single-stage methods}}\\
UnionDet~\cite{kim2020uniondet} & \xmark & $17.58$ & $11.72$ & $19.33$\\
\rowB DIRV~\cite{fang2021dirv} & \xmark & $21.78$ & $16.38$ & $23.39$\\
PPDM-Hourglass~\cite{liao2020ppdm} & \xmark & $21.94$ & $13.97$ & $24.32$\\
\rowB HOI-Transformer~\cite{zou2021end} & \xmark & $23.46$ & $16.91$ & $25.41$\\
GGNet~\cite{zhong2021glance} & \xmark & $23.47$ & $16.48$ & $25.60$\\
\rowB HOTR~\cite{kim2021hotr} & \xmark & $25.10$ & $17.34$ & $27.42$\\
QPIC~\cite{tamura2021qpic} & \xmark & $29.07$ & $21.85$ & $31.23$\\
\rowB CDN~\cite{zhang2021mining} & \xmark & $31.44$ & $27.39$ & $32.64$\\
RLIP~\cite{yuan2022rlip} & VG$^\dagger$ & $32.84$ & $26.85$ & $34.63$\\
\rowB GEN-VLKT~\cite{liao2022gen} & CLIP$^\dagger$ & $\bm{33.75}$ & $\bm{29.25}$ & $\bm{35.10}$\\
\midrule
\rowA \multicolumn{5}{l}{\textit{Bottom-up methods}}\\
InteractNet~\cite{gkioxari2018detecting} & \xmark & $9.94$ & $7.16$ & $10.77$\\
\rowB GPNN~\cite{qi2018learning} & \xmark & $13.11$ & $9.34$ & $14.23$\\
iCAN~\cite{gao2018ican} & \xmark & $14.84$ & $10.45$ & $16.15$\\
\rowB No-Frills~\cite{gupta2019no} & Pose~\cite{cao2017realtime} & $17.18$ & $12.17$ & $18.68$\\
PMFNet~\cite{wan2019pose} & Pose~\cite{lin2014microsoft} & $17.46$ & $15.65$ & $18.00$\\
\rowB CHGNet~\cite{wang2020contextual} & \xmark & $17.57$ & $16.85$ & $17.78$\\
DRG~\cite{gao2020drg} & Text~\cite{mikolov2013efficient} & $19.26$ & $17.74$ & $19.71$\\
\rowB IP-Net~\cite{wang2020learning} & \xmark & $19.56$ & $12.79$ & $21.58$\\
VSGNet~\cite{ulutan2020vsgnet} & \xmark & $19.80$ & $16.05$ & $20.91$\\
\rowB FCMNet~\cite{liu2020amplifying} & Text~\cite{mikolov2013efficient} & $20.41$ & $17.34$ & $21.56$\\
ACP~\cite{kim2020detecting} & Text~\cite{mikolov2013efficient} & $20.59$ & $15.92$ & $21.98$\\
\rowB PD-Net~\cite{zhong2020polysemy} & Text~\cite{mikolov2013efficient} & $20.81$ & $15.90$ & $22.28$\\
DJ-RN~\cite{li2020detailed} & Pose~\cite{cao2017realtime,pavlakos2019expressive} & $21.34$ & $18.53$ & $22.18$\\
\rowB IDN~\cite{li2020hoi} & \xmark & $23.36$ & $22.47$ & $23.63$\\
ATL~\cite{hou2021affordance} & \xmark & $23.81$ & $17.43$ & $25.72$\\
\hline
\rowB \textbf{\OURS} (CLIP: ViT-B/32) & CLIP$^\dagger$ & $29.93$ & $22.94$ & $32.02$\\
\textbf{\OURS} (CLIP: ViT-B/16) & CLIP$^\dagger$ & $31.88$ & $23.04$ & $34.52$\\
\rowB \textbf{\OURS} (CLIP: ViT-L/14) & CLIP$^\dagger$ & $37.41$ & $28.90$ & $39.95$\\
\textbf{\OURS} (LiT: ViT-B/32) & LiT$^\dagger$ & $29.38$ & $23.64$ & $31.09$\\
\rowB \textbf{\OURS} (LiT: R26+B/1) & LiT$^\dagger$ & $33.18$ & $24.78$ & $35.69$\\
\textbf{\OURS} (LiT: ViT-H/14) & LiT$^\dagger$ & $\bm{38.07}$ & $\bm{31.65}$ & $\bm{39.99}$\\
\bottomrule
\end{tabular}}
\end{center}
\caption{\textbf{System-level comparison on the HICO-DET test set.} We report the Mean Average Precision (mAP) under the \textit{Default} setting~\cite{chao2018learning} containing the \textit{Full} (F), \textit{Rare} (R), and \textit{Non-Rare} (N) sets. $^\dagger$ denotes training supervisions obtained from the model pre-training stage. Best performances are highlighted in bold.}
\label{tbl:hico_sota}
\end{table}

\subsection{Human-Object Interaction Detection}
\label{sec:experiments:hoi}

\p{Comparisons to the state of the art} Our framework is compared to both bottom-up and single-stage methods on HICO-DET. We do not include other studies about data augmentation~\cite{park2022consistency,zhong2022towards} and knowledge transfer~\cite{liu2022interactiveness,wu2022mining}, which have different research targets but are complementary to our method. As illustrated in Table~\ref{tbl:hico_sota}, our smallest model (ViT-B/32) matches the state-of-the-art performance, and consistent improvements can be achieved when we scale up the model. Our largest model with a ViT-H/14 backbone outperforms the previous best method by $4.32$ mAP, setting the new state of the art. More importantly, compared to conventional bottom-up approaches, we improve them by $14.26$ mAP, a $60\%$ relative improvement. We hypothesize that these improvements result from (1) image-level pre-training by leveraging VLMs and (2) our bottom-up design, which makes it possible to utilize more object detection datasets. Please refer to the supplementary material for results on V-COCO, where we made similar conclusions.

\begin{figure}[t]
\begin{center}
\includegraphics[width=\linewidth]{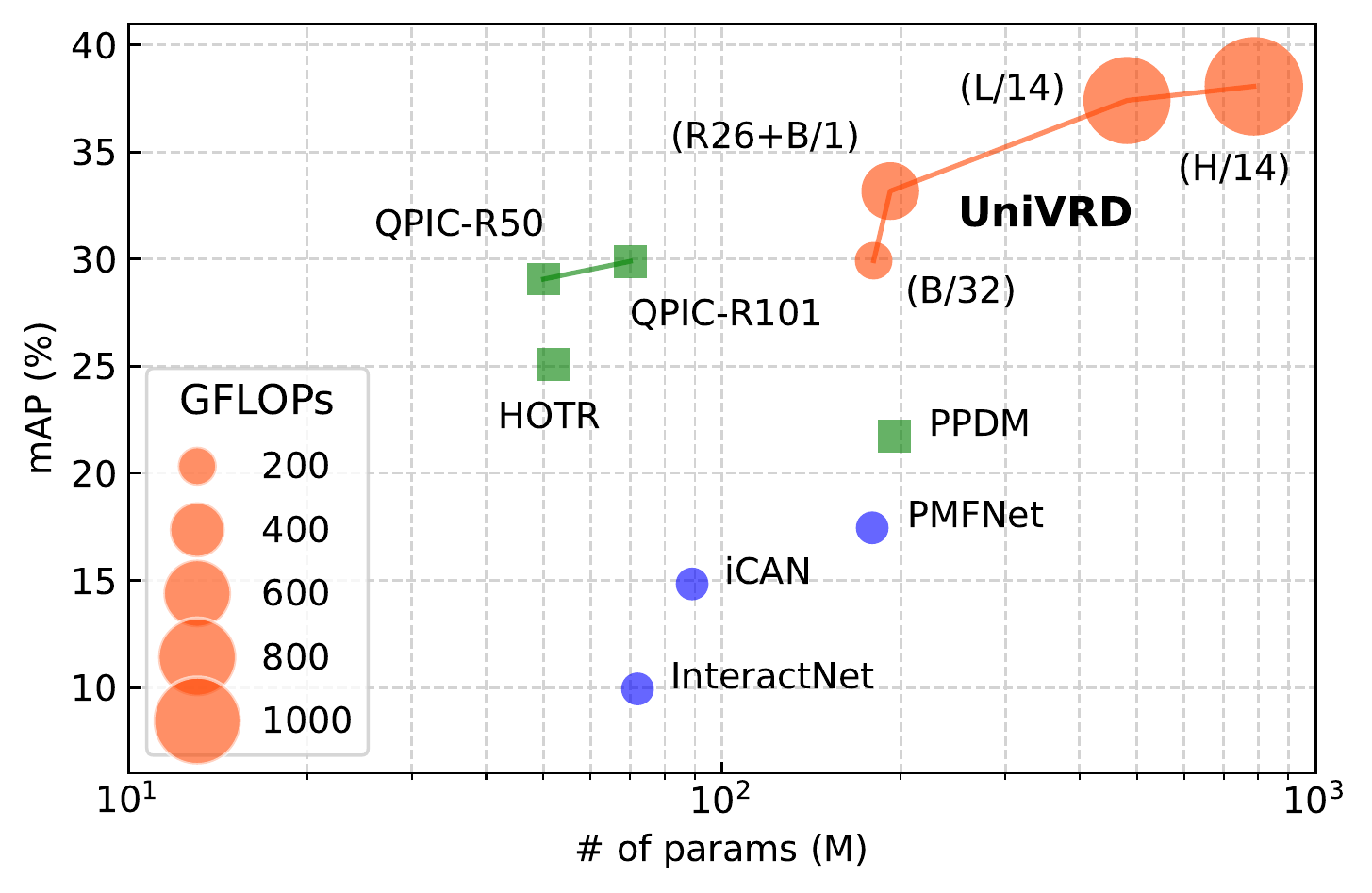}
\end{center}
\vspace{-1.6mm}
\caption{\textbf{Model scale vs.\ performance analysis for HOI detection on the HICO-DET test set.} All circles stand for bottom-up methods. One-stage approaches are marked in green squares. Orange circles represent our models with different backbones and their marker sizes stand for GFLOPs (see the legend).}
\label{fig:scales}
\end{figure}

\begin{table}[t]
\begin{center}
\resizebox{0.98\linewidth}{!}{
\begin{tabular}{lr}
\toprule
Ablation & $\text{mAP}_\text{F}$\\
\midrule
\textit{Full approach} & $29.93$\\
\midrule
\rowA \multicolumn{2}{l}{\textit{Data augmentation}}\\
\textit{(1)} w/o CLIP prompts for the object detector & $-1.85$\\
\textit{(2)} Use CLIP prompts for VRD (at training/inference) & $0.00$\\
\textit{(3)} Use CLIP prompt ensemble for VRD (at inference) & $0.01$\\
\textit{(4)} Use no random crop & $-1.33$\\
\textit{(5)} Use no random horizontal flip & $-1.20$\\
\textit{(6)} Use no mosaics augmentation & $-1.76$\\
\midrule
\rowA \multicolumn{2}{l}{\textit{Training and inference strategy}}\\
\textit{(7)} Use one-stage training schedule & $-5.05$\\
\textit{(8)} Fine-tune the object detector in the second stage & $-1.03$\\
\textit{(9)} Fine-tune the text encoder in the second stage & $-0.92$\\
\textit{(10)} Use no PNMS at all (at inference) & $-2.01$\\
\textit{(11)} Use vanilla PNMS~\cite{zhang2021mining} (at inference) & $-1.24$\\
\midrule
\rowA \multicolumn{2}{l}{\textit{Object detection datasets}}\\
\textit{(12)} w/o Objects365~\cite{shao2019objects365} for object detection & $-3.96$\\
\textit{(13)} w/o COCO~\cite{lin2014microsoft} for object detection & $-2.67$\\
\textit{(14)} w/o VG~\cite{krishna2017visual} for object detection & $-3.16$\\
\bottomrule
\end{tabular}}
\end{center}
\caption{\textbf{Ablation study of the main methodological improvements.} For simplicity, difference in mAP to the \textit{full approach} is shown. All ablations are carried out for the \OURS (CLIP) model with the ViT-B/32 backbone on the HICO-DET dataset.}
\label{tbl:ablation}
\end{table}

\p{Ablation study} Figure~\ref{fig:scales} studies the model scalability (including the parameter numbers and GFLOPs) for HOI detection. A remarkable improvement can be seen when the backbone is switched from ViT-B/32 to R26+B/1 (a hybrid architecture) with only a slight increase in model size and GFLOPs. Further, we can still obtain an increase of $0.7$ mAP when changing ViT-L/14 to ViT-H/14 (our largest model), demonstrating the strong scalability of \OURS.

We provide a detailed ablation study on our method in Table~\ref{tbl:ablation}, which identifies important factors affecting the performance. Especially from \textit{(1)-(6)}, we show how different text prompting and image augmentation techniques influence the model performance. From \textit{(7)-(9)}, we find using a cascade training paradigm substantially boosts the results, while freezing the object detector and text encoder avoids overfitting. This is because accurate object locations and well-aligned image-text embeddings can stabilize decoder optimization. However, allowing fine-tuning these modules leads to performance boosts when we train unified models due to the availability of more training data (see the supplementary material for more details). Additionally, we can see from \textit{(10)-(11)} that performing per-class PNMS also leads to a favorable performance gain than vanilla PNMS~\cite{zhang2021mining}. To further show whether incorporating diverse object detection datasets can benefit HOI detection, we experiment with training our model by removing one object detection dataset each time in \textit{(12)-(14)}. Unsurprisingly, we observe notable performance drops, suggesting the importance of training with diverse object detection datasets.

\subsection{Scene Graph Generation}
\label{sec:experiments:sgg}

\begin{table}[t]
\begin{center}
\resizebox{.98\linewidth}{!}{
\begin{tabular}{lcc}
\toprule
Model & mR@$50$ & mR@$100$\\
\midrule
\rowA \multicolumn{3}{l}{\textit{Specific methods}}\\
KERN~\cite{chen2019knowledge} & $6.4$ & $7.3$\\
\rowB GBNet~\cite{zareian2020bridging} & $7.1$ & $8.5$\\
PCPL~\cite{yan2020pcpl} & $9.5$ & $11.7$\\
\rowB BGNN~\cite{li2021bipartite} & $10.7$ & $12.6$\\
DT2-ACBS~\cite{desai2021learning} & $\bm{22.0}$ & $\bm{24.4}$\\
\midrule
\rowA \multicolumn{3}{l}{\textit{General methods}}\\
GPS-Net~\cite{lin2020gps} & $5.9$ & $7.1$\\
\rowB \celG{GPS-Net~\cite{lin2020gps} w/ Resampling~\cite{li2021bipartite}} & \celG{$7.4$} & \celG{$9.5$}\\
\celG{GPS-Net~\cite{lin2020gps} w/ IETrans + Rwt~\cite{zhang2022fine}} & \celG{$16.2$} & \celG{$18.8$}\\
\rowB Motif~\cite{zellers2018neural} & $6.7$ & $7.7$\\
\celG{Motif~\cite{zellers2018neural} w/ TDE~\cite{tang2020unbiased}} & \celG{$8.2$} & \celG{$9.8$}\\
\rowB \celG{Motif~\cite{zellers2018neural} w/ CogTree~\cite{yu2020cogtree}} & \celG{$10.4$} & \celG{$11.8$}\\
\celG{Motif~\cite{zellers2018neural} w/ DLFE~\cite{chiou2021recovering}} & \celG{$11.7$} & \celG{$13.8$}\\
\rowB \celG{Motif~\cite{zellers2018neural} w/ IETrans + Rwt~\cite{zhang2022fine}} & \celG{$15.5$} & \celG{$18.0$}\\
VCTree~\cite{tang2019learning} & $6.7$ & $8.0$\\
\rowB \celG{VCTree~\cite{tang2019learning} w/ TDE~\cite{tang2020unbiased}} & \celG{$9.3$} & \celG{$11.1$}\\
\celG{VCTree~\cite{tang2019learning} w/ CogTree~\cite{yu2020cogtree}} & \celG{$10.4$} & \celG{$12.1$}\\
\rowB \celG{VCTree~\cite{tang2019learning} w/ DLFE~\cite{chiou2021recovering}} & \celG{$11.8$} & \celG{$13.8$}\\
\celG{VCTree~\cite{tang2019learning} w/ IETrans + Rwt~\cite{zhang2022fine}} & \celG{$12.0$} & \celG{$14.9$}\\
\rowB SG-Transformer~\cite{yu2020cogtree} & $7.7$ & $9.0$\\
\celG{SG-Transformer~\cite{yu2020cogtree} w/ CogTree~\cite{yu2020cogtree}} & \celG{$11.1$} & \celG{$12.7$}\\
\rowB \celG{SG-Transformer~\cite{yu2020cogtree} w/ IETrans + Rwt~\cite{zhang2022fine}} & \celG{$16.2$} & \celG{$18.8$}\\
\hline
AS-Net$^\dagger$~\cite{chen2021reformulating} & $6.1$ & $7.2$\\
\rowB HOTR$^\dagger$~\cite{kim2021hotr} & $9.4$ & $12.0$\\
\hline
\textbf{\OURS} (CLIP: ViT-B/32) & $9.6$ & $12.1$\\
\rowB \textbf{\OURS} (CLIP: ViT-B/16) & $10.9$ & $13.2$\\
\textbf{\OURS} (CLIP: ViT-L/14) & $\bm{12.6}$ & $\bm{14.5}$\\
\bottomrule
\end{tabular}}
\end{center}
\caption{\textbf{Performance on the Visual Genome test set.} We adopt the metric mean Recall@$K$ (mR@$K$). $^\dagger$ denotes methods designed for HOI detection and the results are reproduced with their open-source code. Best performances are highlighted in bold.}
\label{tbl:vg_sota}
\end{table}

\begin{table*}[t]
\begin{center}
\resizebox{\linewidth}{!}{
\begin{tabular}{lcccccccccc}
\toprule
& & & \multicolumn{3}{c}{HICO-DET} & \multicolumn{2}{c}{V-COCO} & \multicolumn{3}{c}{Visual Genome}\\
\cmidrule(lr){4-6}\cmidrule(lr){7-8}\cmidrule(lr){9-11}
Model & Backbone & GFLOPs & $\text{mAP}_\text{F}$ & $\text{mAP}_\text{R}$ & $\text{mAP}_\text{N}$ & $\text{mAP}_\text{S\#1}$ & $\text{mAP}_\text{S\#2}$ & mR@$50$ & mR@$100$ & mAP\\
\midrule
\textbf{\OURS} (CLIP) & ViT-B/32 & $185$ & $29.47$\numD{0.46} & $22.93$\numD{0.01} & $31.42$\numD{0.60} & $39.48$\numU{5.21} & $40.81$\numU{5.73} & \multicolumn{1}{r}{$9.61$\numD{0.01}} & $12.04$\numD{0.08} & $8.64$\numU{0.16}\\
\rowB \textbf{\OURS} (CLIP) & ViT-B/16 & $436$ & $32.80$\numU{0.92} & $24.64$\numU{1.60} & $35.24$\numU{0.72} & $41.57$\numU{5.32} & $43.39$\numU{6.49} & $10.58$\numD{0.32} & $12.77$\numD{0.47} & $8.21$\numD{0.76}\\
\textbf{\OURS} (CLIP) & ViT-L/14 & $992$ & $38.61$\numU{1.20} & $33.39$\numU{4.49} & $40.16$\numU{0.21} & $45.19$\numU{5.39} & $46.52$\numU{6.30} & $12.55$\numD{0.08} & $14.48$\numD{0.10} & $9.85$\numD{0.12}\\
\bottomrule
\end{tabular}}
\end{center}
\caption{\textbf{Performance of our unified detectors (trained with HICO-DET, V-COCO, and VG) at different scales on multiple datasets.} We also report the performance differences (\textbf{\textcolor{blue}{$\uparrow$~improvement}} and \textbf{\textcolor{red}{$\downarrow$~drop}}) compared with their dataset-specific counterparts.}
\label{tbl:unify}
\end{table*}

We categorize the SGG approaches into two categories: (1) general models~\cite{lin2020gps,tang2019learning,yu2020cogtree,zellers2018neural} referring to methods that can be equipped with model-agnostic baselines~\cite{chiou2021recovering,li2021bipartite,tang2020unbiased,yu2020cogtree,zhang2022fine} in a plug-and-play manner; (2) specific models~\cite{chen2019knowledge,desai2021learning,li2021bipartite,yan2020pcpl,zareian2020bridging} indicating dedicated designed models with strong performance. Table~\ref{tbl:vg_sota} shows that our model achieves competitive performance compared with the state of the art. To see how models transfer between HOI detection and SGG, we also report the results of two HOI detectors~\cite{chen2021reformulating,kim2021hotr} on VG. We find that our model still outperforms them by a large margin. Note that the training data of SGG are highly biased and contain more noisy annotations than HOI detection, which make it hard for a model to achieve strong results without specific designs. Our model can be directly integrated with model-agnostic baselines (colored in gray) to further boost the results, since our design makes no specific assumptions on the targeted VRD tasks. We also note that DT2-ACBS~\cite{desai2021learning} achieves the best performance among specific methods, since it proposed specific architectures and sampling strategies to handle the long-tailed issue in VG, which is orthogonal to our method.

\subsection{Towards Unified VRD Models}
\label{sec:experiments:unified}

One major advantage brought by unifying label spaces is the capability of training models over a pool of various datasets, but prior studies~\cite{lambert2020mseg,wang2019towards} observe performance drops in single unified models. We next quantitatively compare our unified model with dataset-specific ones to see performance changes, especially, when models are scaled up.

We report the results of our model at different scales in Table~\ref{tbl:unify}, where we make three important observations. First, both $\text{mAP}_\text{S\#1}$ and $\text{mAP}_\text{S\#2}$ on V-COCO are significantly improved (with more than $5.5$ mAP on average) by our unified models, because the training set is extremely small in V-COCO (around $5,000$ images). This improvements verify the effectiveness of our unified training schema: training models with other datasets simultaneously reduces overfitting and encourages models to transfer learnt knowledge to small-scale datasets like V-COCO. Second, our unified models perform as well as their dataset-specific counterparts in mAP on HICO-DET and VG. Notably, the improvements further increase on HICO-DET along with the growth of model sizes, suggesting the necessity for scaling up to large models when we train unified detectors across multiple datasets. Third, all unified models have slight performance drops in recalls on the VG dataset. This is potentially because training on more HOI datasets (\ie, HICO-DET and V-COCO) yields higher accuracy in the HOI domain, but loses ground on non-HOI relationships in VG.

\begin{figure*}[t]
\begin{center}
\includegraphics[width=\linewidth]{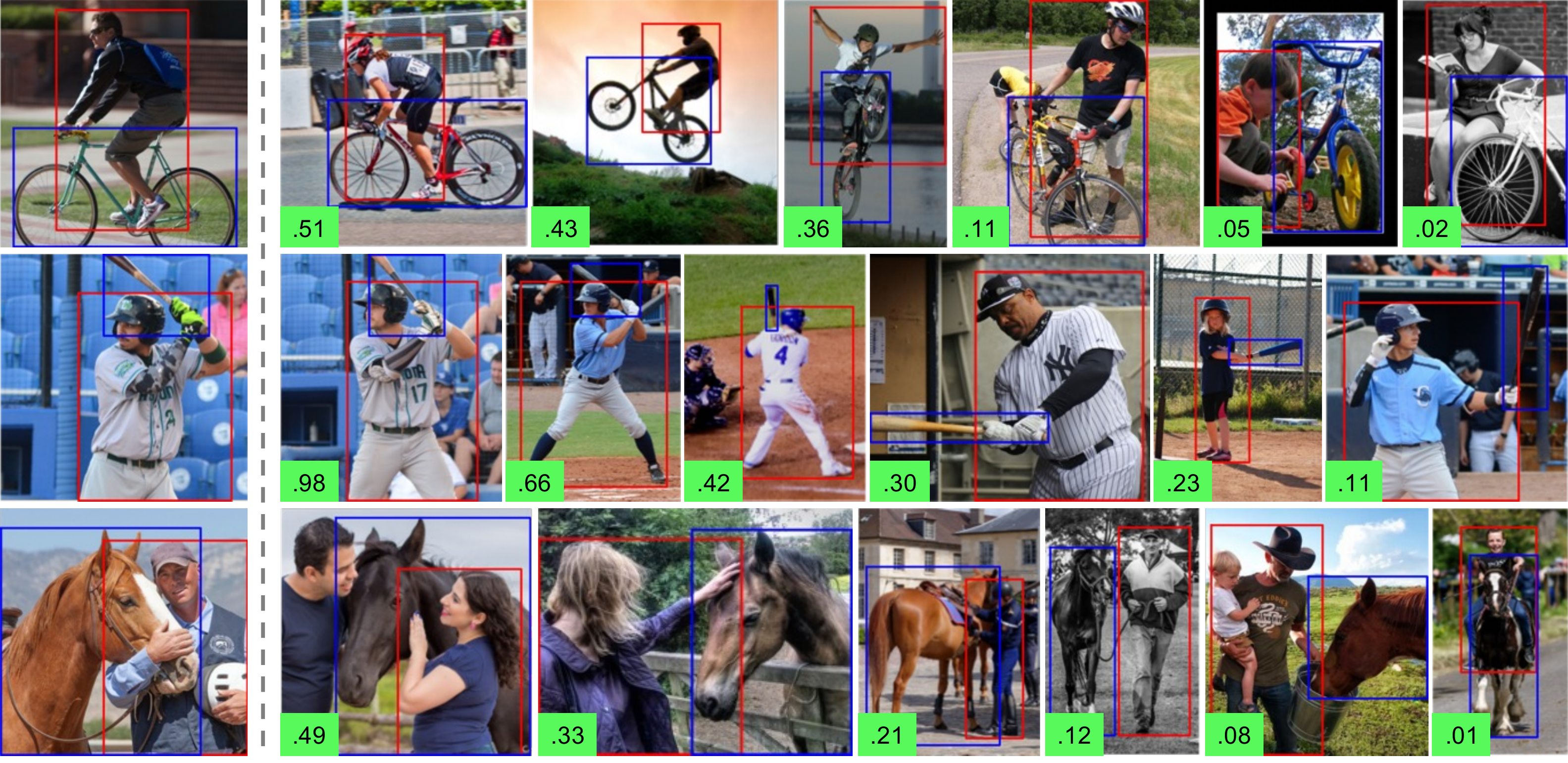}
\end{center}
\vspace{-1.6mm}
\caption{\textbf{Visual examples of image-conditioned relationship retrievals using relation embeddings.} The leftmost images show the queries, and corresponding retrievals from the HICO-DET test set based on their relation embeddings are shown on the right. We mark subjects in red, objects in blue, and similarity scores in green boxes.}
\label{fig:visual}
\end{figure*}

\p{Image-based relation retrieval} Our model can perform image-conditioned detection by simply replacing text query embeddings with image-derived ones. Here we show visual illustrations for this use case. To get the image query embedding, we first run inference on the query image and select the top-$1$ prediction. We then use its image embedding as query on the test images. Figure~\ref{fig:visual} visually shows examples of image-conditioned relationship retrievals for the given image queries ranked by similarity scores. By using image query embeddings, our model enables retrieval of relationships which would be hard to describe in text.

\p{Limitations} Our algorithm currently does not specially handle extremely biased or long-tailed relationship categories, which may widely exist in the wild. Using auxiliary priors from data (through transferring~\cite{zhang2022fine} or resampling~\cite{li2021bipartite}) may further improve the performance on SGG datasets (\eg, VG). Our formulation currently does not explicitly formulate relationship hierarchies, where both objects and predicates are reasoned at a single shot. We leave incorporating more powerful VQA-VLMs (\eg, PaLI~\cite{chen2022pali}) to model such hierarchies as exciting future work.
 \section{Conclusion}
\label{sec:conclusion}

We presented a bottom-up recipe for training a single unified visual relationship detection (VRD) model across multiple datasets based on vision and language models. Our resulting detector shows competitive performances under both dataset-specific and unified configurations on two VRD tasks: human-object interaction detection and scene graph generation. For the first time, we show scaling up to large models can benefit unified models on VRD tasks. We hope our model serves as a strong baseline approach towards general VRD systems. 
{\small
\bibliographystyle{ieee_fullname}
\bibliography{egbib}
}

\clearpage
\begin{appendices}
\section{Implementation Details}

\subsection{Architecture}

\p{Vision and language model (VLM)} We adopt the pre-trained CLIP~\cite{radford2021learning} and LiT~\cite{zhai2022lit}, which both provide open-source implementations and pre-trained checkpoints. We experiment with CLIP backbones of ViT-B/32, ViT-B/16, and ViT-L/14, and LiT backbones of ViT-B/32, R26+B/1, and ViT-H/14. These backbones follow the nomenclature from~\cite{dosovitskiy2020image} for model size, patch size, and Transformer \vs hybrid architectures. For example, B/32 refers to ViT-Base with patch size $32$, while R26+B/1 refers to a hybrid ResNet-26 plus ViT-Base with stride $1$. We use the pre-trained CLIP weights provided by the original paper and obtain the pre-trained LiT weights from OWL-ViT~\cite{minderer2022simple}.

\p{Object detector} To adapt VLMs for object detection, we remove token pooling and add detection heads, which contain one linear layer for producing classification embeddings and the other two-layer feed-forward network for box prediction. Following the design of OWL-ViT~\cite{minderer2022simple}, we add a bias to the predicted box coordinates so that each box is by default centered on the image patch that corresponds to the token from which this box is predicted when arranging the token sequence as a 2D grid. Although the stochastic depth regularisation~\cite{huang2016deep} (\ie, droplayer) is not applied during VLM pre-training, we add it to image encoders during fine-tuning, which reduces model overfitting. In addition, we merge the class token into other feature map tokens by multiplying it with them, and append layer norm to the output of CLIP models, following the practice in~\cite{minderer2022simple}.

\p{Relationship decoder} We use Perceiver Resampler proposed in Flamingo~\cite{alayrac2022flamingo} as our decoder backbone. It contains three Transformer layers with eight attention heads. Note that we employ ReLU instead of Squared ReLU used in Flamingo to simplify the design. We set the number of input relation queries to $100$ for all datasets. Each Transformer layer is implemented following the post-normalization design~\cite{vaswani2017attention}, which requires a linear warm-up learning rate schedule ($1,000$ warm-up steps) for model training.

\begin{table*}[t]
\begin{center}
\resizebox{0.96\linewidth}{!}{
\begin{tabular}{lcccccccc}
\toprule
Model & Backbone & \# of steps & Batch size & Learning rate & Droplayer rate & Image size & Dataset proportions & Frozen text\\
\midrule
\OURS (CLIP) & ViT-B/32 & $140,000$ & $256$ & $5 \times 10^{-5}$ & $0.2$ & $768$ & $0.1/0.5/0.2/0.2$ & \cmark\\
\rowB \OURS (CLIP) & ViT-B/16 & $140,000$ & $256$ & $5 \times 10^{-5}$ & $0.2$ & $768$ & $0.1/0.5/0.2/0.2$ & \cmark\\
\OURS (CLIP) & ViT-L/14 & $70,000$ & $256$ & $2 \times 10^{-5}$ & $0.2$ & $672$ & $0.1/0.5/0.2/0.2$ & \cmark\\
\hline
\rowB \OURS (LiT) & ViT-B/32 & $140,000$ & $256$ & $2 \times 10^{-4}$ & $0.2$ & $768$ & $0.1/0.7/0.1/0.1$ & \cmark\\
\OURS (LiT) & R26+B/1 & $140,000$ & $256$ & $2 \times 10^{-4}$ & $0.2$ & $768$ & $0.1/0.7/0.1/0.1$ & \cmark\\
\rowB \OURS (LiT) & ViT-H/14 & $70,000$ & $256$ & $5 \times 10^{-5}$ & $0.2$ & $480$ & $0.1/0.7/0.1/0.1$ & \cmark\\
\bottomrule
\end{tabular}}
\end{center}
\caption{\textbf{List of hyper-parameters used for training our object detector.} The mix probabilities of the COCO~\cite{lin2014microsoft}, Objects365~\cite{shao2019objects365}, HICO-DET~\cite{chao2018learning}, and Visual Genome~\cite{krishna2017visual} datasets within each batch are shown in dataset proportions. Note that we only apply stochastic depth regularisation~\cite{huang2016deep} (\ie, droplayer) to image encoders, as text encoders are frozen.}
\label{tbl:supp:obj_det}
\end{table*}

\begin{table*}[t]
\begin{center}
\resizebox{\linewidth}{!}{
\begin{tabular}{lcccccccc}
\toprule
Model & Backbone & \# of steps & Batch size & Learning rate & Droplayer rate & Image size & Dataset proportions & Frozen text\\
\midrule
\rowA \multicolumn{9}{l}{\textit{Dataset-specific models}}\\
\OURS (CLIP) & ViT-B/32 & $140,000$ & $64$ & $1 \times 10^{-4}$ & $0.0$ & $768$ & - & \cmark\\
\rowB \OURS (CLIP) & ViT-B/16 & $140,000$ & $64$ & $1 \times 10^{-4}$ & $0.0$ & $768$ & - & \cmark\\
\OURS (CLIP) & ViT-L/14 & $140,000$ & $64$ & $1 \times 10^{-4}$ & $0.0$ & $672$ & - & \cmark\\
\rowA \multicolumn{9}{l}{\textit{Unified models}}\\
\OURS (CLIP) & ViT-B/32 & $140,000$ & $256$ & $1 \times 10^{-4} / 2 \times 10^{-6}$ & $0.2$ & $768$ & $0.5/0.1/0.4$ & \xmark\\
\rowB \OURS (CLIP) & ViT-B/16 & $140,000$ & $256$ & $1 \times 10^{-4} / 2 \times 10^{-6}$ & $0.2$ & $768$ & $0.5/0.1/0.4$ & \xmark\\
\OURS (CLIP) & ViT-L/14 & $140,000$ & $256$ & $1 \times 10^{-4} / 2 \times 10^{-6}$ & $0.2$ & $672$ & $0.5/0.1/0.4$ & \xmark\\
\bottomrule
\end{tabular}}
\end{center}
\caption{\textbf{List of hyper-parameters used for training our visual relationship decoder.} Where two numbers are given for the learning rate, the first is for the visual relationship decoder
and the second for the rest of the whole model. The mix probabilities of the HICO-DET~\cite{chao2018learning}, V-COCO~\cite{gupta2015visual}, and Visual Genome~\cite{krishna2017visual} datasets within each batch are shown in dataset proportions. Note that we only apply stochastic depth regularisation~\cite{huang2016deep} (\ie, droplayer) to image encoders.}
\label{tbl:supp:vrd_det}
\end{table*}

\subsection{Training Object Detector}

We list the hyper-parameters used for training our object detector in Table~\ref{tbl:supp:obj_det}. Its training procedure follows~\cite{minderer2022simple}, except that we make two major modifications. First, we freeze the text encoder of a pre-trained VLM. This is because we would like to keep the embedding space of a pre-trained text encoder so that its discriminative capability is still sufficient for encoding relationship triplets, preventing it from forgetting issues. Second, we apply the stochastic depth regularisation~\cite{huang2016deep} (\ie, droplayer) to image encoders when using LiT~\cite{zhai2022lit} as the pre-trained VLM; otherwise, performance drops will be observed. When training CLIP models, we mix the COCO~\cite{lin2014microsoft}, Objects365~\cite{shao2019objects365}, HICO-DET~\cite{chao2018learning}, and Visual Genome~\cite{krishna2017visual} datasets randomly in each batch with probabilities of $0.1$, $0.5$, $0.2$, and $0.2$, respectively; when training LiT models, we use probabilities of $0.1$, $0.7$, $0.1$, and $0.1$, respectively. This is due to the fact that LiT models suffer from overfitting if a larger probability is applied to HICO-DET or Visual Genome.

\subsection{Training Relationship Decoder}

In the experiments, we introduce two configurations for training our relationship decoders: dataset-specific models and unified models. They use different training setups which are shown in Table~\ref{tbl:supp:vrd_det}.

\p{Dataset-specific models} To reduce model overfitting, we fix both the text encoder and object detector when training dataset-specific models. We train all models in this configuration with $140,000$ steps, the learning rate of $1 \times 10^{-4}$, and batch size $64$. When training dataset-specific models on the V-COCO dataset, we observe serious overfitting problems as its training set is extremely small (less than $5,000$ images). Therefore, we early stop the model training, where we use at most $20,000$ training steps, while keeping other hyper-parameters unchanged.

\p{Unified models} When training unified models, we mix HICO-DET, V-COCO, and Visual Genome randomly in each batch with probabilities of $0.5$, $0.1$, and $0.4$, respectively. We find that further increasing the mix ratio of training data from Visual Genome will lead to model overfitting on this dataset. We enlarge the batch size to $256$ and set the learning rate of both the text encoder and object detector to $2 \times 10^{-6}$. We have also tried the same optimization setup for training dataset-specific models, but it will lead to performance drops. This suggests the benefits of enlarging the batch size and unfreezing pre-trained models when we train unified models across multiple datasets.

\section{Additional Results}

\subsection{Mosaics Image Augmentation}

Table~\ref{tbl:supp:mosaics} shows our results on HICO-DET~\cite{chao2018learning} when different mosaics configurations are employed for training the visual relationship decoder. To highlight the performance differences, we report the results without using per-class PNMS. We can find that using only $1 \times 1$ single images is clearly worse than including larger mosaics (\ie, smaller
mosaic tiles), and the model achieves the best performance with the inclusion of $3 \times 3$ mosaics.

\begin{table}[t]
\begin{center}
\resizebox{0.84\linewidth}{!}{
\begin{tabular}{cccccc}
\toprule
\multicolumn{3}{c}{Mosaics ratio} & \multicolumn{3}{c}{Default ($\%$)}\\
\cmidrule(lr){1-3}\cmidrule(lr){4-6}
$1 \times 1$ & $2 \times 2$ & $3 \times 3$ & $\text{mAP}_\text{F}$ & $\text{mAP}_\text{R}$ & $\text{mAP}_\text{N}$\\
\midrule
$1.0$ & $0.0$ & $0.0$ & $25.84$ & $20.09$ & $27.56$\\
$0.6$ & $0.4$ & $0.0$ & $27.06$ & $19.66$ & $29.27$\\
$0.4$ & $0.3$ & $0.3$ & $27.92$ & $20.98$ & $30.00$\\
\bottomrule
\end{tabular}}
\end{center}
\caption{\textbf{Performance comparison when different mosaics ratios are utilized for image augmentation.} We report the results of \OURS (CLIP) using the ViT-B/32 backbone on the HICO-DET test set without performing per-class PNMS.}
\label{tbl:supp:mosaics}
\end{table}

\subsection{Ablation on Training Unified Models}

\begin{table}[t]
\begin{center}
\resizebox{0.96\linewidth}{!}{
\begin{tabular}{lr}
\toprule
Ablation & $\text{mAP}_\text{F}$\\
\midrule
\textit{Unified baseline} & $29.47$\\
\midrule
\textit{(1)} Use one-stage training schedule & $-4.97$\\
\textit{(2)} Freeze the object detector in the second stage & $-1.21$\\
\textit{(3)} Freeze the text encoder in the second stage & $-0.83$\\
\bottomrule
\end{tabular}}
\end{center}
\caption{\textbf{Ablation study of the main methodological improvements for training unified models.} For simplicity, difference in mAP to the \textit{unified baseline} is shown. All ablations are carried out for the \OURS (CLIP: ViT-B/32) model on HICO-DET.}
\label{tbl:ablation_unified}
\end{table}

We identify the top three important factors affecting the performance of our unified models in Table~\ref{tbl:ablation_unified}. We can find that using a cascade training paradigm still leads to a substantial performance boost, which is consistent with our observations on training dataset-specific models. In contrast, freezing either the object detector or the text encoder when training the relationship decoder causes performance drops. This is because of the fact that our proposed unified VRD framework makes it possible for us to train models with a larger amount of data across multiple datasets at the same time, mitigating the model overfitting issue. Hence, we are able to train models with larger learning capabilities (with more model parameters to be fine-tuned).

\subsection{Results with Different Backbones}

To conduct a fair comparison on the backbone, we show results of different models using the same ResNet-50 backbone in Table~\ref{tbl:supp:R50}. We can observe that our method achieves competitive performances on both HICO-DET and VG.

\begin{table}[t]
\begin{center}
\resizebox{0.82\linewidth}{!}{
\begin{tabular}{cccc}
\toprule
Method & HOTR & QPIC & \textbf{UniVRD}\\
\midrule
HICO-DET (mAP$_\text{F}$) & $25.1$ & $29.1$ & $\bm{29.7}$\\
Visual Genome (mR@$50$) & $9.4$ & - & $\bm{9.6}$\\
\bottomrule
\end{tabular}}
\end{center}
\caption{\textbf{Results with the ResNet-50 backbone on HICO-DET and VG.} Our model is initialized from CLIP~\cite{radford2021learning}.}
\label{tbl:supp:R50}
\end{table}

\subsection{HOI Detection on V-COCO}

\begin{table}[t]
\begin{center}
\resizebox{0.88\linewidth}{!}{
\begin{tabular}{lccc}
\toprule
Model & Extra-sup. & $\text{AP}^\text{S\#1}_\text{role}$ & $\text{AP}^\text{S\#2}_\text{role}$\\
\midrule
\rowA \multicolumn{4}{l}{\textit{Single-stage methods}}\\
UnionDet~\cite{kim2020uniondet} & \xmark & $47.5$ & $56.2$\\
\rowB HOI-Transformer~\cite{zou2021end} & \xmark & $52.9$ & -\\
GGNet~\cite{zhong2021glance} & \xmark & $54.7$ & -\\
\rowB HOTR~\cite{kim2021hotr} & \xmark & $55.2$ & $64.4$\\
DIRV~\cite{fang2021dirv} & \xmark & $56.1$ & -\\
\rowB QPIC~\cite{tamura2021qpic} & \xmark & $58.8$ & $61.0$\\
CDN~\cite{zhang2021mining} & \xmark & $61.7$ & $63.8$\\
\rowB RLIP~\cite{yuan2022rlip} & VG$^\dagger$ & $61.9$ & $64.2$\\
GEN-VLKT~\cite{liao2022gen} & CLIP$^\dagger$ & $\bm{62.4}$ & $\bm{64.5}$\\
\midrule
\rowA \multicolumn{4}{l}{\textit{Bottom-up methods}}\\
InteractNet~\cite{gkioxari2018detecting} & \xmark & $40.0$ & -\\
\rowB GPNN~\cite{qi2018learning} & \xmark & $44.0$ & -\\
iCAN~\cite{gao2018ican} & \xmark & $45.3$ & $52.4$\\
\rowB TIN~\cite{li2019transferable} & \xmark & $47.8$ & $54.2$\\
VCL~\cite{hou2020visual} & \xmark & $48.3$ & -\\
\rowB DRG~\cite{gao2020drg} & Text~\cite{mikolov2013efficient} & $51.0$ & -\\
IP-Net~\cite{wang2020learning} & \xmark & $51.0$ & -\\
\rowB VSGNet~\cite{ulutan2020vsgnet} & \xmark & $51.8$ & $57.0$\\
PMFNet~\cite{wan2019pose} & Pose~\cite{lin2014microsoft} & $52.0$ & -\\
\rowB PD-Net~\cite{zhong2020polysemy} & Text~\cite{mikolov2013efficient} & $52.6$ & -\\
CHGNet~\cite{wang2020contextual} & \xmark & $52.7$ & -\\
\rowB FCMNet~\cite{liu2020amplifying} & Text~\cite{mikolov2013efficient} & $53.1$ & -\\
ACP~\cite{kim2020detecting} & Text~\cite{mikolov2013efficient} & $53.2$ & -\\
\rowB IDN~\cite{li2020hoi} & \xmark & $53.3$ & $60.3$\\
\hline
\textbf{\OURS} (CLIP: ViT-B/32) & CLIP$^\dagger$ & $59.9$ & $62.7$\\
\rowB \textbf{\OURS} (CLIP: ViT-B/16) & CLIP$^\dagger$ & $62.3$ & $64.8$\\
\textbf{\OURS} (CLIP: ViT-L/14) & CLIP$^\dagger$ & $65.1$ & $66.3$\\
\rowB \textbf{\OURS} (LiT: ViT-B/32) & LiT$^\dagger$ & $59.4$ & $62.2$\\
\textbf{\OURS} (LiT: R26+B/1) & LiT$^\dagger$ & $62.6$ & $65.1$\\
\rowB \textbf{\OURS} (LiT: ViT-H/14) & LiT$^\dagger$ & $\bm{65.8}$ & $\bm{66.9}$\\
\bottomrule
\end{tabular}}
\end{center}
\caption{\textbf{System-level comparison on V-COCO.} $^\dagger$ denotes training supervisions obtained from the model pre-training stage. Best performances are highlighted in bold.}
\label{tbl:vcoco_sota}
\end{table}

In this section, we provide additional results on the V-COCO dataset~\cite{gupta2015visual}. The metric of role AP is used for evaluation: a detection is correct if the location of the agent (\ie, both subjects and objects) and each role (\ie, predicate classes) is correct (correctness is measured using bounding box overlap as is standard). In V-COCO, there are a number of HOI categories which are defined with no object labels. To deal with this situation, we evaluate the model performance in two different scenarios following the official evaluation scheme of V-COCO. In Scenario 1 ($\text{AP}^\text{S\#1}_\text{role}$), detectors are required to report cases in which there is no object,
while in Scenario 2 ($\text{AP}^\text{S\#2}_\text{role}$), we just ignore the prediction of an object bounding box in these cases.

To deal with the long-tail class distribution in V-COCO, we use the dynamic re-weighting~\cite{zhang2021mining} during model training. To handle HOI categories which do not contain objects, we conduct the following modifications to let them be compatible with the proposed framework. First, for each sample, their subject annotations are employed as pseudo ground-truth objects to be predicted by our model. Second, we use the prompt template `a $\langle$subject$\rangle$ $\langle$predicate$\rangle$-ing' for HOI categories including transitive verbs and the prompt template `a $\langle$subject$\rangle$ $\langle$predicate$\rangle$-ing something' for those containing intransitive verbs.

We compared the proposed method with both bottom-up and single-stage methods. As illustrated in Table~\ref{tbl:vcoco_sota}, we can find that (1) we are able to achieve the state-of-the-art performance; (2) our model outperforms other bottom-up  approaches by a significant margin; (3) further improvements can be obtained when we scale up the model. These observations are consistent with our results on HICO-DET, which re-confirm the effectiveness of the proposed method. \end{appendices}

\end{document}